\documentclass{article}
\usepackage{ijcai24}

\usepackage{times}
\usepackage{soul}
\usepackage{url}
\usepackage[hidelinks]{hyperref}
\usepackage[utf8]{inputenc}
\usepackage[small]{caption}
\usepackage{graphicx}
\usepackage{amsmath}
\usepackage{amsthm}
\usepackage{booktabs}
\usepackage{algorithm}
\usepackage[switch]{lineno}

\usepackage{times}
\usepackage{latexsym}
\usepackage[T1]{fontenc}
\usepackage[utf8]{inputenc}
\usepackage{microtype}
\usepackage{xspace}

\newcommand{\ours}{\textsc{FactCHD}}
\newcommand{\ourmodel}{\textsc{Truth-Triangulator}}
\usepackage{dirtytalk}
\usepackage{colortbl}
\usepackage{booktabs}
\usepackage{transparent}
\usepackage{xcolor,soul}
\usepackage{booktabs}
\usepackage{enumitem}
\usepackage{etoolbox}
\usepackage{colortbl}
\usepackage{transparent}

\usepackage{amssymb}
\usepackage{url}

\usepackage{graphicx,calc}
\usepackage{wrapfig}
\usepackage{color,soul}
\usepackage{enumitem}
\usepackage{multirow}
\usepackage{bm}
\usepackage{booktabs}
\usepackage{arydshln}
\usepackage{cleveref}
\usepackage{hyperref}
\usepackage{url}
\usepackage{algorithm}
\usepackage{algorithmicx}
\usepackage[noend]{algpseudocode}
\usepackage{enumitem}

\def\emojitruth{\raisebox{-0.55ex}{\includegraphics[width=1.3em]{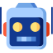}}}
\def\emojiseeker{\raisebox{-0.55ex}{\includegraphics[width=1.3em]{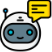}}}
\def\emojijudge{\raisebox{-0.55ex}{\includegraphics[width=1.3em]{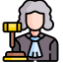}}}

\usepackage{color}
\usepackage{xcolor}
\usepackage{amsfonts}
\usepackage{latexsym}
\usepackage{url}
\usepackage{wrapfig}
\usepackage{comment}
\usepackage{booktabs}
\usepackage{microtype}
\usepackage{xspace}
\usepackage{booktabs}
\usepackage{multirow}
\usepackage{dcolumn}
\usepackage{hhline}
\usepackage{mdframed}
\usepackage{enumitem}
\usepackage{xcolor}
\usepackage{csquotes}
\newcolumntype{d}[1]{D{.}{.}{#1}}
\usepackage{subcaption}
\usepackage{makecell}
\usepackage{color,soul}
\usepackage{algorithm}
\usepackage[noend]{algpseudocode}
\usepackage{xcolor,colortbl}
\usepackage{xspace}
\usepackage{mdframed}
\usepackage[outline]{contour}
\usepackage{hyperref}

\makeatletter
\newcommand\footnoteref[1]{\protected@xdef\@thefnmark{\ref{#1}}\@footnotemark}
\makeatother

\definecolor{mygray}{gray}{.9}
\newcommand{\GG}{\cellcolor{mygray}}

\usepackage{xcolor}
\definecolor{gold}{RGB}{205,133,63}
\definecolor{fGreen}{RGB}{34,139,34}
\definecolor{tOrange}{RGB}{255,165,0}
\definecolor{tBlue}{RGB}{135,206,250}
\definecolor{tPink}{RGB}{255,204,204}
\definecolor{tGreen}{RGB}{210,200,225}
\definecolor{tPurple}{RGB}{250,200,225}
\definecolor{tGold}{RGB}{255,215,150}
\usepackage{tabularx}

\newcommand{\colordmark}{\textcolor{my_green}{\ding{52}}}

\newcolumntype{b}{X}
\newcolumntype{m}{>{\hsize=.6\hsize}X}
\definecolor{aliceblue}{RGB}{178, 217, 245}

\definecolor{babyblue}{RGB}{217, 239, 251}

\usepackage{float}
\usepackage{enumitem}
\usepackage{tikz}
\usepackage{tcolorbox}
\usepackage{siunitx}
\usepackage{pifont}% 

\newcommand{\1}{\uppercase\expandafter{\romannumeral1}}
\newcommand{\2}{\uppercase\expandafter{\romannumeral2}}

\newcommand{\daugshifted}{\raisebox{0.5\depth}{$\uparrow$}}
\newcommand{\dashifted}{\raisebox{0.5\depth}{\tiny$\downarrow$}}

\newcommand{\uashifted}{\raisebox{0.9\depth}{\tiny$\uparrow$}}
\newcommand{\da}[1]{{\scriptsize\hlprimarytab{\dashifted{#1}}}}
\newcommand{\ua}[1]{{\scriptsize\hlsecondarytab{\uashifted{#1}}}}

\newcommand{\daulg}[1]{{\hlsecondarytab{\daugshifted{#1}}}}

\definecolor{c3}{cmyk}{0.3081,0,0.7209,0.3255} 

\newtcbox{\hlprimarytab}{on line, rounded corners, box align=base, colback=c3!10,colframe=white,size=fbox,arc=3pt, before upper=\strut, top=-2pt, bottom=-4pt, left=-2pt, right=-2pt, boxrule=0pt}
\newtcbox{\hlsecondarytab}{on line, box align=base, colback=red!10,colframe=white,size=fbox,arc=3pt, before upper=\strut, top=-2pt, bottom=-4pt, left=-2pt, right=-2pt, boxrule=0pt}

\definecolor{my_green}{RGB}{51,102,0}
\definecolor{my_red}{RGB}{204, 0, 0}

\newcommand{\colorimark}{\textcolor{my_red}{\ding{55}}}

\title{FactCHD: Benchmarking Fact-Conflicting Hallucination Detection}

\author{
Xiang Chen$^{1,4}$,
Duanzheng Song$^{2}$,
Honghao Gui$^{1,4}$,
Chenxi Wang$^{2,4}$,
Ningyu Zhang$^{2,4}$\footnotemark[1],\\
Yong Jiang$^{3}$,
Fei Huang$^{3}$,
Chengfei Lv$^{3}$,
Dan Zhang$^{2}$,
Huajun Chen$^{1,4}$\thanks{~~Corresponding author.}\\ 
\affiliations
 $^1$College of Computer Science and Technology, Zhejiang University\\ 
  $^2$School of Software Technology, Zhejiang University\\
   $^3$Alibaba Group \\
  $^4$ZJU-Ant Group Joint Research Center for Knowledge Graphs, Zhejiang University\\ 
 \emails
  {\{xiang\_chen, sdz, guihonghao, sunnywcx, zhangningyu, dan.zhang, huajunsir\}@zju.edu.cn}
  \\
  \{yongjiang.jy, f.huang, chengfei.lcf\}@alibaba-inc.com \\
}

\begin{document}

\maketitle

\begin{abstract}

Despite their impressive generative capabilities, LLMs are hindered by fact-conflicting hallucinations in real-world applications. The accurate identification of hallucinations in texts generated by LLMs, especially in complex inferential scenarios, is a relatively unexplored area. To address this gap, we present \textbf{\ours}, a dedicated benchmark designed for the detection of fact-conflicting hallucinations from LLMs. \textbf{\ours} features a diverse dataset that spans various factuality patterns, including vanilla, multi-hop, comparison, and set operation. A distinctive element of \textbf{\ours} is its integration of fact-based evidence chains, significantly enhancing the depth of evaluating the detectors' explanations. Experiments on different LLMs expose the shortcomings of current approaches in detecting factual errors accurately. Furthermore, we introduce \textbf{\ourmodel} that synthesizes reflective considerations by tool-enhanced ChatGPT and LoRA-tuning based on Llama2, aiming to yield more credible detection through the amalgamation of predictive results and evidence.
\end{abstract}
% \url{https://github.com/zjunlp/FactCHD}

\section{Introduction}

Large Language Models (LLMs)~\cite{LLM-survey}  are susceptible to generating text that, while seemingly credible, can be factually inaccurate or vague, leading to the spread of misinformation online~\cite{Know2023,hallucination-survey,TRUSTGPT}. This issue, referred to as \textit{fact-conflicting hallucination}~\cite{suvery-Siren}, arises from the incorporation of incorrect or obsolete knowledge into the models' parameters and from the models' inherent limitations in complex cognitive ability. These shortcomings constrain LLMs' deployment in critical domains like finance, healthcare, and law, and amplify the propagation of erroneous information. 
Therefore, it is crucial to effectively detect fact-conflicting hallucinations for mitigating or editing them~\cite{edit_survey}.
\begin{figure}[htb!] %H为当前位置，!htb为忽略美学标准，htbp为浮动图形
\centering %图片居中
\includegraphics[width=0.4\textwidth]{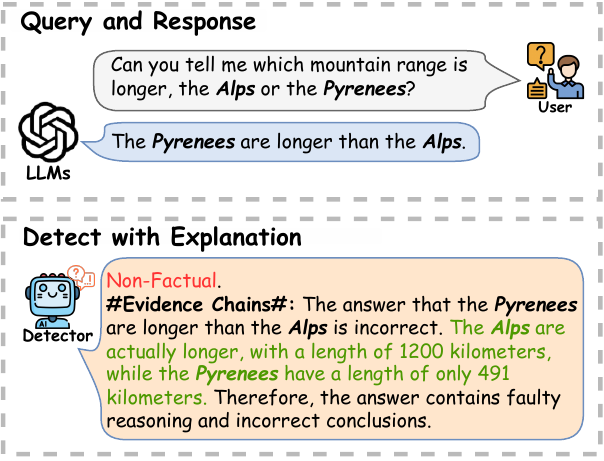} %
% 插入图片，[]中设置图片大小，{}中是图片文件名
\caption{Illustration of fact-conflicting hallucination detection example from {\ours}, where the green part represents factual explanation core (body part) in the chain of evidence.} 
\label{fig:intro}
\end{figure}

However, traditional fact verification tasks~\cite{scifact,wadden-etal-2020-fact} are not suitable for LLM-based ``\textsc{Query-Response}'' data. Moreover, existing hallucination evaluation benchmarks~\cite{HaluEval,factor}, predominantly centering on vanilla facts and textual content, lack in-depth exploration of complex operations among facts, thus rendering their coverage against fact-conflicting hallucinations suboptimal. 
To bridge this gap, we introduce an inherently rigorous\cite{kg-conflict}, yet authentic, task scenario: \emph{fact-conflicting hallucination} detection, devoid of explicit claims or evidence. 
As shown in Figure~\ref{fig:intro}, when confronted with a query and its generated response, detectors, are impelled to harness both their intrinsic knowledge and external resources, while rendering a factual judgment accompanied by an elucidative explanation.

\begin{table*}[!t]
    \centering
    \small
    \vskip -0.3in
    \vskip 0.1in
    \scalebox{0.85}{
    \begin{tabular}{lccccc}
    \toprule
    \textbf{Datasets} 
    & \textbf{\#  Source} 
    & \textbf{Domains} 
     % & \textbf{Construction} 
    % & \textbf{Format} 
    & \textbf{Factuality Pattern} & \textbf{Evaluation}
    & \textbf{Expandability} \\
    \midrule
    \textsc{Fever}~\cite{thorne-etal-2018-fever} & Text & General  & \textsc{Van.} & \textsc{Consist.}
    & \colorimark \\
    \textsc{Climate-Fever}~\cite{diggelmann2020climate} & Text  & Climate Change   & \textsc{Van.} & \textsc{Consist.} 
    & \colorimark \\
    \textsc{Health-Fever}~\cite{healthfever} & Text & Health  & \textsc{Van.} & \textsc{Consist.} 
    & \colorimark \\
    \textsc{Sci-Fact}~\cite{scifact} & Text &Scientific    &\textsc{Van.} & \textsc{Consist.} 
    & \colorimark \\
    \textsc{CoVERT}~\cite{CoVERT} & Text &  COVID-19  & \textsc{Van.} & \textsc{Consist.} 
     & \colorimark \\
    \textsc{TabFact}~\cite{TabFact} & Table & General   & \textsc{Van.} & \textsc{Consist.} 
    & \colorimark \\
    \textsc{HoVer}~\cite{HoVer} & Text &  General  & \textsc{Mul.} & \textsc{Consist.} 
     & \colorimark \\
    \textsc{FEVEROUS}~\cite{FEVEROUS} & Text+Table &  General & \textsc{Van.} & \textsc{Consist.} 
     & \colorimark \\
    \textsc{HaluEval}~\cite{HaluEval} & Text & General  &\textsc{Van.} & {Hallucination} 
     & \colorimark  \\
    \midrule
    {\ours} (ours) 
    & 
    \textbf{KGs}
    (\begin{minipage}[b]{0.06\columnwidth}
    \centering
    \raisebox{-.3\height}{\includegraphics[width=1.2\linewidth]{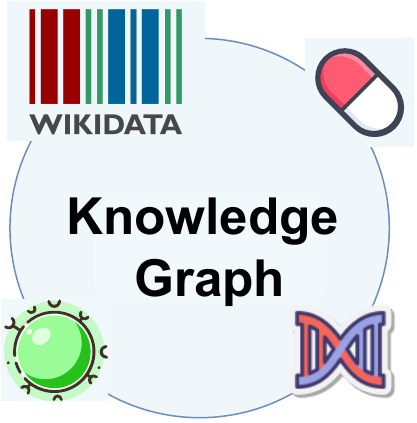}}
\end{minipage})
\& Text
& General\&Vertical   &  \textsc{Van.}\&\textsc{Mul.}\&\textsc{Com.}\&\textsc{Set.} & \textsc{Fact.}+\textsc{Chain.} 
     & \colordmark \\
    \bottomrule
    \end{tabular}
    }
        \caption{
    \small
    Comparison with existing fact-checking datasets.
    Our {\ours} include both general and vertical domains, such as health, COVID-19, climate, science, and medicine (genes, virus and disease).
\textsc{Van.}, \textsc{Mul.}, \textsc{Com.} and \textsc{Set.} are abbreviations for the distinct factuality patterns.
    %We utilize the following abbreviations: \textbf{HG} - human-curated datasets; \textbf{MIX} - datasets containing both human-constructed and machine-generated data.
    %\textbf{Q\&R} - ``\textsc{Query-Response}''  context; \textbf{\textsc{Consist.}} - classification of consistency between claims and given evidence; \textbf{\textsc{Fact.}} - classification of factuality, with labels factual and non-factual; \textbf{\textsc{Chain.}} - chains of evidence.
}
    \vskip -0.1in
    \label{tab:compare}
\end{table*}

% Consistency classification

% factual classification

In paving the way for future strides in hallucination evaluation, we introduce a new benchmark, \textbf{\underline{Fact}}-\textbf{\underline{C}}onflicting \textbf{\underline{H}}allucination \textbf{\underline{D}}etection (\ours), tailored for LLMs and encompassing a variegated array of factuality patterns, including \textbf{Vanilla}, \textbf{Multi-hops}, \textbf{Comparison}, and \textbf{Set-Operation} patterns. For example in Figure 1, querying whether the ``Alps'' or  ``Pyrenees'' are higher, exemplifies a comparison pattern, assessing the relative relationships among facts.
Drawing inspiration from the adage \textit{``to know it and to know the reason why of it''} by \emph{Zhuzi}, {\ours} extends beyond mere ``\textsc{Query-Response}''  labeling of hallucinations, incorporating golden chains of evidence to assess if detectors can provide coherent explanations for factualness judgment.
Acknowledging the challenges of collecting comprehensive data through exhaustive human annotation, we propose a scalable data construction approach that harnesses existing knowledge graphs (KGs) and textual knowledge to create simulated hallucination instances with ChatGPT, verified with human annotation, for the efficient development of {\ours}.

We evaluate the performance of various LLMs (such as Alpaca, Llama2-chat, and ChatGPT) using our {\ours} benchmark across multiple settings: zero-shot, in-context learning, specialized detection tuning, and knowledge enhancement via retrieval/tools. The results indicate that specialized detection tuning and knowledge enhancement notably improve the detection of fact-conflicting hallucinations.
Additionally, we present \textbf{\ourmodel} framework grounded in ``Triangulation'' theory~\cite{triangulation}. This system comprises three roles: ChatGPT, enhanced with tools as the \emph{Truth Seeker}; a detect-specific expert based on Llama2-7B-LoRA as the \emph{Truth Guardian}; and the \emph{Fact Verdict Manager}, which amasses evidence from another role to fortify the reliability and accuracy of the derived conclusions. {\ourmodel} emphasizes the use of cross-referencing generators to astutely evaluate and adjudicate responses with potential factual discrepancies. Key insights are summarized as:
\begin{itemize}
    \item
    We present \ours\footnote{Data is available at \url{https://github.com/zjunlp/FactCHD}.}, a large-scale, multi-domain evaluation benchmark with diverse factual patterns and interpretable evidence chains, setting a new standard for detecting fact-conflicting hallucinations from LLMs.
    
    \item 
 We introduce a scalable data construction strategy leveraging KGs, etc. to efficiently develop a fact-conflicting hallucination dataset, offering stronger applicability due to the authentic, broad domain coverage of KGs.

    \item We devise a triangulation-based framework {\ourmodel} that employs cross-referencing generators for verifying LLM responses. 
    %Empirical results demonstrate the effectiveness of our approach in hallucination detection.

\end{itemize}

\section{Related Work}

\paragraph{Hallucination in LLMs.}

\begin{figure*}[tb] %H为当前位置，!htb为忽略美学标准，htbp为浮动图形
\centering %图片居中
\includegraphics[width=1.0\textwidth]{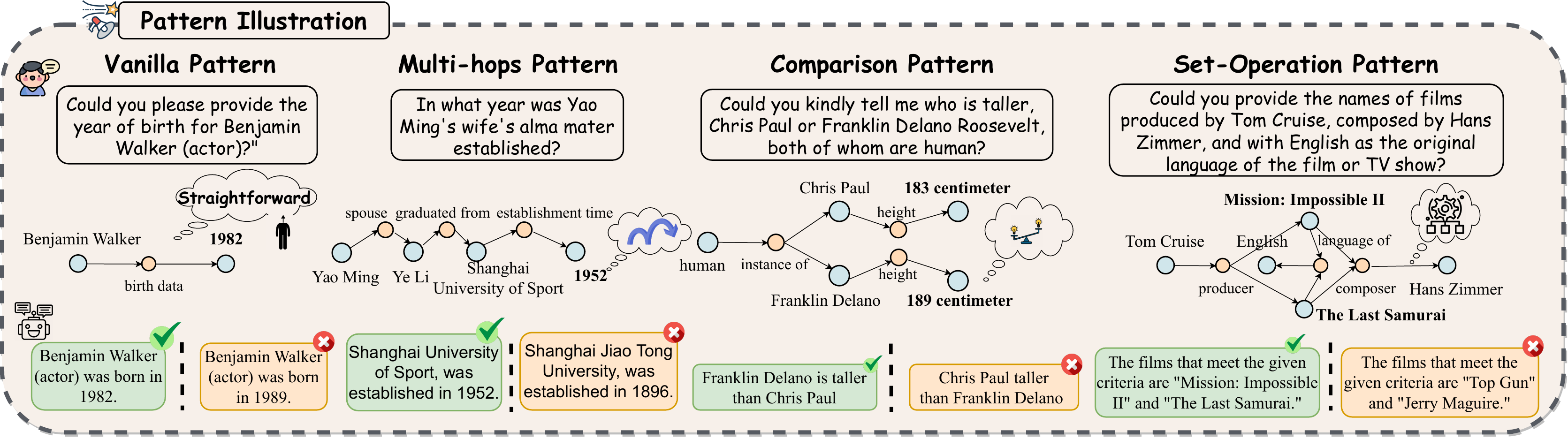} %
% 插入图片，[]中设置图片大小，{}中是图片文件名
  \caption{Overview of the factuality patterns invloved in our \ours.}\label{fig:pattern}
\end{figure*}

\begin{figure}[tb] %H为当前位置，!htb为忽略美学标准，htbp为浮动图形
\centering %图片居中
\includegraphics[width=0.5\textwidth]{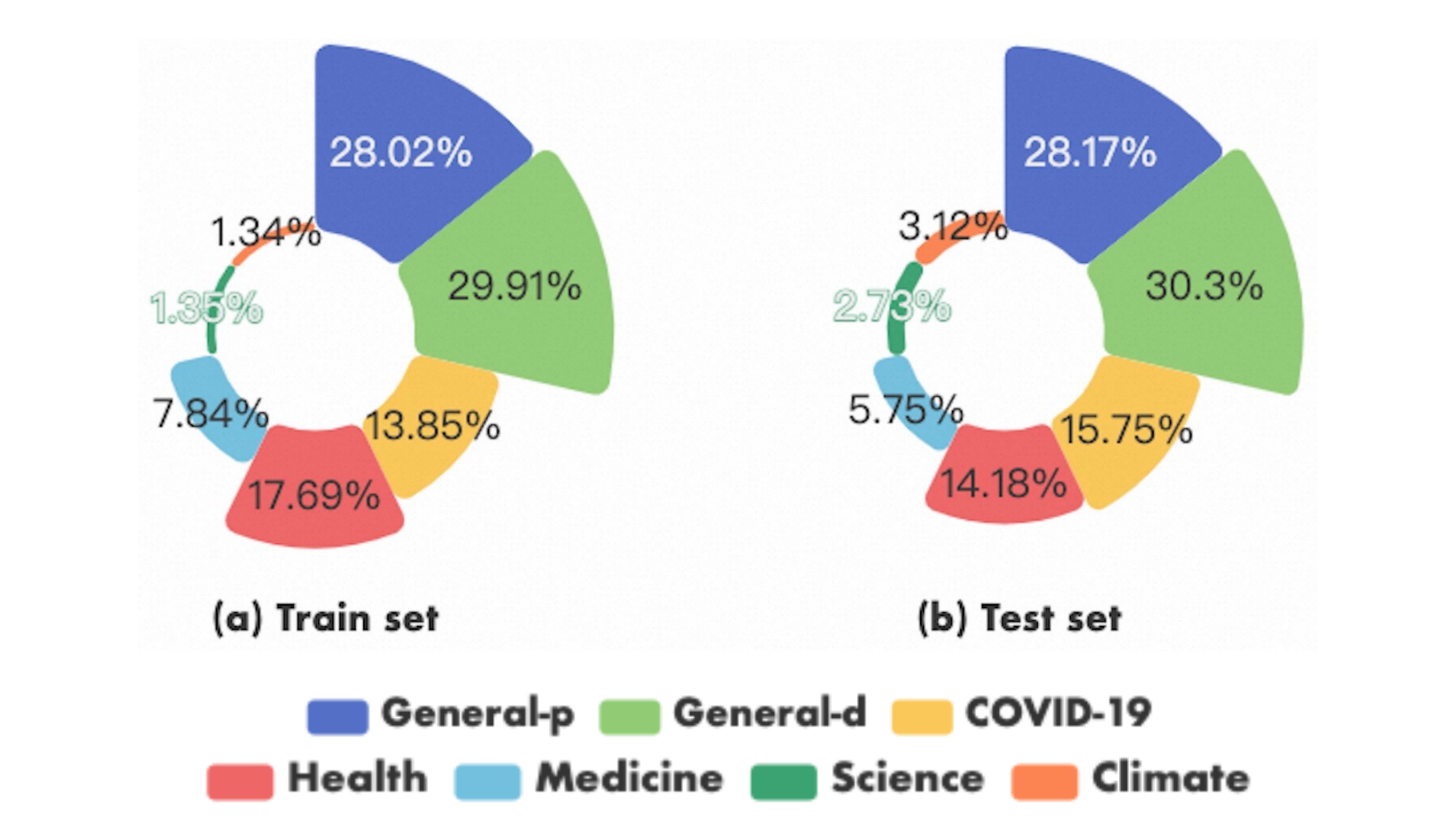} 
\small
\caption{Domain distribution of  \ours, where ``-p'' and ``-d'' denote  domains derived from Wikipedia and Wikidata, respectively.
}
\label{fig:distribution}
\end{figure}

Despite LLMs~\cite{LLM-survey,knowprompt,chatgpt} like ChatGPT demonstrating remarkable understanding and execution of user instructions, they are prone to confidently generating misleading ``hallucinations''~\cite{hallucination-survey,fact_survey}. These hallucinations, as categorized by \cite{suvery-Siren}, can be input-conflicting, context-conflicting, or fact-conflicting—with the latter being especially problematic due to the propagation of inaccurate factual information online.
While previous studies have extensively examined hallucinations within natural language generation (NLG) for various NLP tasks~\cite{shuster2021retrieval,creswell2022faithful,dialogue-hallucination,mallen2022not}, HaluEval~\cite{HaluEval} has emerged as a recent benchmark for assessing LLMs' recognition of such errors. 
Different from HaLuEval which only analyzes ChatGPT's ability to evaluate whether hallucinatory, we specifically focus on the evaluation of fact-conflicting hallucination by constructing an interpretable benchmark that can serve as a public platform for checking the factual errors of context with explanation.

\paragraph{Factuality Detection.}

Our research is also related to prior works on fact verification in NLP tasks~\cite{thorne-etal-2018-fever,wadden-etal-2020-fact,GuptaWLX22,DziriKMZYPR22,keliang23,Rashkin-Measuring,kryscinski2020evaluating,CRAG}, expanding the understanding of factuality beyond binary judgments. The FRANK framework~\cite{pagnoni2021understanding} offers a detailed typology of factual errors, while ~\cite{DhingraFPCDC19} explores lexical entailment in text generation. The FACTOR score by~\cite{factor} evaluates LMs based on the likelihood of factual content.
Existing benchmarks often miss the ``\textsc{Query-Response}''  context of LLMs, focusing solely on accuracy without providing explanatory rationales. Our contribution lies in: (1) presenting ``\textsc{Query-Response}''  formatted data for LLMs; (2) introducing metrics for interpretability in detecting fact-conflicting hallucinations; and (3) validating the effectiveness of {\ourmodel} through cross-referential verification from multiple sources. The comparison with other datasets is detailed in Table~\ref{tab:compare}. 

% Additionally, our emphasis on factuality aids in foundational work for model editing of LLMs~\cite{edit_survey,mao2023editing,li2023unveiling,easyedit}.

% FELM~\cite{chen2023felm} extends actuality evaluation into math and reasoning tasks. 

%  If no inconsistency is
% detected, the summary is consistent. The detection
% framing also allows for models to provide natural
% language explanations when identifying a summary
% as inconsistent, which can be manually verified to
% confirm model reasoning ability, and model failure

% 任务定义，多写公式
% introduction要web一点，段落的开头简述这段话在干啥。
% 论文思考要有深度，围绕case更深度影响，有些类型相对而言容易被查出来。再强调KG在这里能产生什么作用，future work。

\section{Preliminaries}
\label{sec:preliminaries}

% We introduce the task of fact-conflicting hallucination detection and factuality patterns as follows.

\paragraph{Task Formulation.}
Detecting fact-conflicting hallucinations in LLMs involves discerning factual errors in responses to human queries. A comprehensive detector must not only classify responses as factual or non-factual but also provide explanations for its judgments. We define the task as follows:
\textbf{Input:} A question $Q$ paired with an LLM-generated response $R$, which may contain various fact conflicts.
\textbf{Output:} A combined label and explanation sequence $A=[l,e]$, where $l$ is the binary factuality label (\textsc{factual} or \textsc{non-factual}), and $e$ articulates the rationale behind the assigned label.{
We evaluate the quality of $e$ using the $ExpMatch$ metric through the golden evidence chains in {\ours}.}

\paragraph{Factuality Patterns.}
We aim to explore distinct patterns of factual errors in our {\ours}, vividly illustrated in Figure~\ref{fig:pattern}. These include the \emph{(1) vanilla pattern} dealing with factual statements that can be objectively verified using established sources, the \emph{(2) multi-hops pattern} involving the process of concluding by connecting multiple pieces of facts, the \emph{(3) comparison pattern} referring to the act of evaluating and comparing relative worth and relations between different pieces of facts, and the \emph{(4) set-operation pattern} involving manipulating and combining sets of elements using operations to analyze relations between different facts. We generated corresponding ``\textsc{Query-Response}'' examples based on these patterns.

\paragraph{{Knowledge-Driven Factual Foundation for Reliable ``\textsc{Query-Response}'' Generation.}} 

KGs~\cite{tkde/LiangLZTWYDL24} serve as a rich trove of structured entities and relations, ideal for compositional reasoning and anchoring factual data. Alongside this, textual knowledge is critical for nuanced inference beyond basic facts. 
Our research focuses on collecting existing knowledge and integrating it into prompts as a factual foundation for ``\textsc{Query-Respons}'' and chain of evidence generation, as outlined below:
(1) We use $438$ widespread relations from Wikidata~\cite{vrandevcic2014wikidata} and PrimeKG~\cite{chandak2023building} to create varied subgraphs via $K$-hop walks, forming the knowledge base for generating ``\textsc{Query-Respons}'' examples with multi-hop reasoning, fact comparison, and set operation patterns.
(2) We utilize text knowledge from datasets such as FEVER~\cite{thorne-etal-2018-fever}, Climate-Fever~\cite{diggelmann2020climate}, Health-Fever~\cite{healthfever}, COVID-FACT~\cite{covid-fact}, and SCIFACT~\cite{scifact} for generating examples with vanilla pattern.

% We filter only the \textsc{factual} (\textit{supported}) and \textsc{non-factual} (\textit{refuted}) instances along with their evidence, using ChatGPT (\textit{GPT-3.5-turbo}) to assess the dataset's claims, focusing on samples where the model exhibits difficulty in avoiding incorrect responses.

\section{ {\ours} Benchmark Construction}
\label{sec:construction}

% Based on the aforementioned definitions, we construct {\ours}, which comprises a comprehensive array of training samples, enriched with an additional 6,960 meticulously curated samples for evaluating fact-conflicting hallucinations generated by LLMs. Our dataset ensures an equitable balance between \textsc{factual} or \textsc{non-factual} categories, providing a sturdy platform for evaluation. 
% The data statistic and the distribution of {\ours} across various domains are illustrated in Table~\ref{tab:statistic_pattern} and Figure~\ref{fig:distribution}.
% We proceed to introduce the design principles of the fact-conflicting hallucination detection benchmark in the following parts.

Building on the aforementioned preliminaries, we develop {\ours}, a dataset containing a wealth of training instances and an additional 6,960 carefully selected samples for evaluating fact-conflicting hallucinations from LLMs. Our dataset maintains a balanced representation of \textsc{factual} and \textsc{non-factual} categories, offering a robust framework for assessment.
The statistics and domain distribution of {\ours} are depicted in Table~\ref{tab:statistic_pattern} and Figure~\ref{fig:distribution}.
Next, we outline the design principles of our benchmark as follows.

% \subsection{Diverse Source Data Collection}
% \label{sec:data_collect}
 % We have found a very limited number of datasets available with ``\textsc{Query-Response}'' format and factual evidence. 
%  \subsection{Query and Response Data Simulation}
% \label{sec:data_collect}
% In order to automatically generate factuality evaluation data with interpretable chains of evidence, we collect data from two types of sources: KG-based knowledge sources and textual knowledge sources as factual pieces  to  generate high-quality chains of evidence for explanations. Specifically, our {\ours} is extracted from following  sources: 

\begin{table}[!tb]
% \renewcommand\arraystretch{1.1}
% \setlength\tabcolsep{1pt}
	% \footnotesize
    \small
	\centering
	% \setlength\tabcolsep{2.8pt}
  % \resizebox{0.5\textwidth}{!}{%
  \scalebox{0.85}{
	\begin{tabular}{l c c c c c}
		\toprule
		  \textbf{Split} & \textbf{\#Sample} & \textsc{Van.} & \textsc{Multi.}& \textsc{Comp.} & \textsc{Set-Op.} \\
		\midrule
		%NAR NAT       & 24.04         & 3.88      & 20.32  \\
		\textbf{Train} & 51,383  & 31,986  & 8,209 & 5,691 & 5,497 \\
            \textbf{Test}  & 6,960  & 4,451  & 1,013 & 706 & 790 \\

		% \midrule
		\bottomrule
	\end{tabular}
 }\small
	\caption{Data Statistic of our {\ours}}
        % \vspace{-1cm}
	\label{tab:statistic_pattern}
\end{table}

% \begin{table}[!tb]
% % \renewcommand\arraystretch{1.1}
% % \setlength\tabcolsep{1pt}
% 	% \footnotesize
% \caption{Distribution of our proposed  {\ours} over each pattern of factuality.}	\centering
% 	% \setlength\tabcolsep{2.8pt}
%   % \resizebox{0.5\textwidth}{!}{%
%   \scalebox{0.95}{
% 	\begin{tabular}{l rr }
% 		\toprule
% 		  \textbf{Split} 
%     &  \textbf{Train}
%     &   \textbf{Test}   \\
% 		\midrule
% 		%NAR NAT       & 24.04         & 3.88      & 20.32  \\
% 		\textbf{\#Sample} & 51,383  & 6,960  \\
%   \midrule
%         \textbf{Conventional} & 31,986  & 4,451  \\
%         \textbf{Multi-hops } & 8,209
%          & 1,013\\
%          \textbf{Comparison} & 5,691
%          & 706 \\
%          \textbf{Set-Operation} & 5,497 & 790
%         \\

% 		% \midrule
% 		\bottomrule
% 	\end{tabular}
%  }
%         % \vspace{-0.3cm}
% 	\label{tab:statistic_pattern}
% \end{table}

% \emph{(1) conventional pattern} deals with factual statements that can be objectively verified using established sources and are typically more straightforwardly identified. 
% We primarily construct this pattern of data using textual knowledge sources, which posse straightforward evidence.
% \emph{(2) multi-hops pattern} denotes the process of drawing conclusions by connecting multiple pieces of facts.
% \emph{(3) comparison pattern} refers to the act of evaluating and comparing the relative worth, and relationships between different pieces of facts. 
% \emph{(4) set-operation pattern} involves manipulating and combining sets of elements using operations to analyze relationships between different facts.

 \subsection{Collect Realistic Data  as Demonstration}
 \label{sec:collect}

Subsequently, adhering to the defined factuality patterns, we manually craft corresponding queries and utilize open-source  LLMs (such as ChatGLM, Alpaca, and Vicuna) to generate authentic responses. By manually annotating these responses for hallucinations, we acquire ``\textsc{Query-Response}''  examples annotated with hallucination presence. Our objective is to curate a robust dataset of hallucination cases that closely emulate real-world examples, providing a demonstrative foundation and establishing standards for prompt refinement to ensure alignment between generated hallucinations and these demonstrations.

\begin{figure*}[tb] %H为当前位置，!htb为忽略美学标准，htbp为浮动图形
\centering %图片居中
\includegraphics[width=1.0\textwidth]{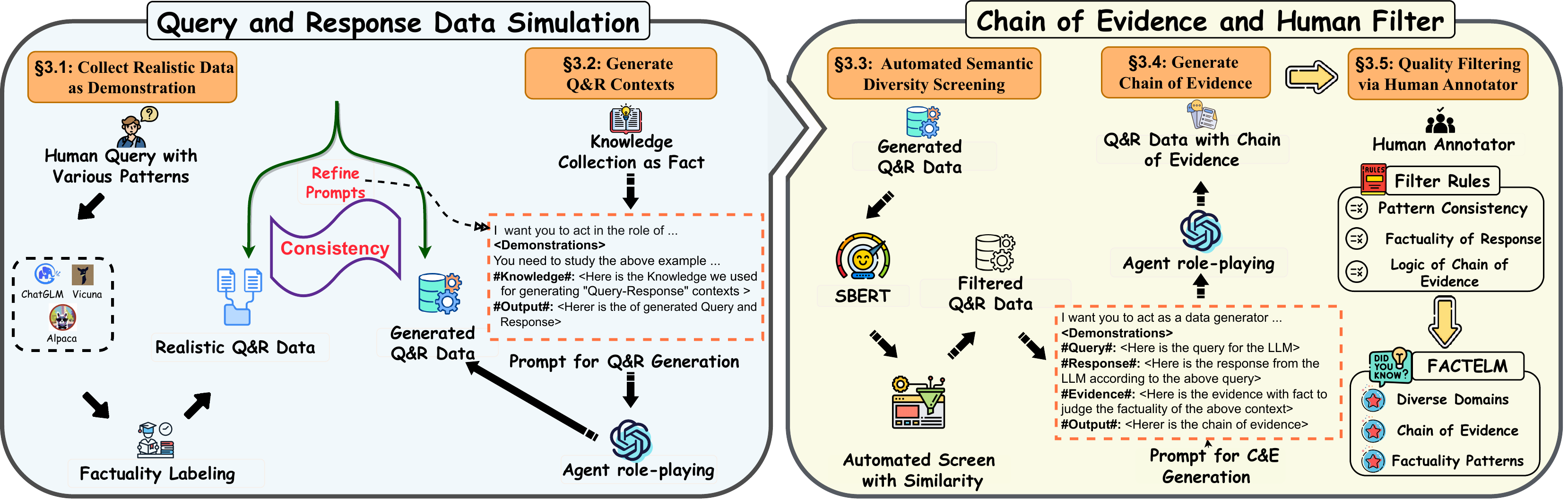} %
% 插入图片，[]中设置图片大小，{}中是图片文件名
  \caption{Overview of the construction process of \ours.}\label{fig:overreview}
\end{figure*}

 \subsection{Generate ``\textsc{Query-Response}'' Contexts} 
\label{sec:generate_qa}

\paragraph{Generate ``\textsc{Query-Response}'' based on Knowledge with ChatGPT.}

 We incorporate the knowledge collected on distinct factuality patterns into customized prompts, specifying whether to generate factual or non-factual responses. This guides ChatGPT in producing ``\textsc{Query-Respons}'' instances across various factuality categories with golden  labels.
% Detailed prompts are in Table~\ref{tab:instruction11} through ~\ref{tab:instruction14} of  ~\S\ref{app:bench_construct}.

 \paragraph{Refine Prompts with Consistency.}

In the initial phase of generating ``\textsc{Query-Response}'' instances, we assess the consistency of five samples per pattern against demonstrations, using majority consensus. We subjectively evaluate each context's adherence to the style and form of realistic demonstrations, employing iterative refinements of prompts to ensure that the generated data exhibits realistic patterns. We aim for an at least 95\% consistency rate before scaling up the generation of ``\textsc{Query-Response}''  instances.

% Aiming to improve the diversity of generated ``\textsc{Query-Response}'' contexts, we employ Sentence-BERT (SBERT)\cite{sbert} in an automated screening process, calculating semantic similarity matrices within contexts. This facilitates the identification and exclusion of highly similar samples, safeguarding dataset diversity. After the screening process, a total of 1,538 samples from the training set and 632 samples from the test set were removed, ensuring the final aggregate of samples is diverse. This meticulous elimination of semantically analogous entries ensures our collections exhibit varied and non-redundant queries and responses, enhancing the benchmark's utility for evaluating diverse instances of fact-conflicting hallucination.

 \subsection{Automated Semantic Diversity Screening}

To increase ``\textsc{Query-Response}'' context diversity, we employ Sentence-BERT (SBERT)\cite{sbert} to automatically compute semantic similarity matrices, filtering out near-duplicate samples to preserve dataset variety.  During filtration, we removed 1,542 training and 832 test samples, guaranteeing a diverse final dataset. This careful pruning of semantically similar entries promotes a varied collection of queries and responses, bolstering the benchmark's effectiveness for assessing diverse instances.

 \subsection{Generate Chain of Evidence}
Our benchmark evaluates the detectors' ability to not only identify hallucinations but also to provide effective explanations. It necessitates ChatGPT's generation of coherent golden evidence chains grounded in factual knowledge for substantiating judgments. Utilizing subgraph or textual facts outlined in \S\cref{sec:preliminaries}, along with the previously generated ``\textsc{Query-Response}'' pairs, ChatGPT delivers thorough justifications for labels assigned to ``\textsc{Query-Response}''  contexts. 
%{Detailed prompts are presented in Tables~\ref{tab:instruction2}} 
These golden evidence chains are critical for assessing the explanatory validity of hallucination detectors.

 \subsection{Quality Filtering via Human Annotator}
We craft filter rules to ensure pattern consistency, response factuality, and logical evidence chains for quality control in the annotation. Several educated annotators are uniformly trained and utilize both their expertise and search tools for rigorous sample vetting. 
To minimize subjectivity, we organized them into groups of three, incorporating a voting mechanism for the evaluation of data.
Simultaneously judged mismatches by annotators led to sample discarding, resulting in final removal counts of 565 and 258 samples from the training and test sets, respectively.
We involve Fleiss's Kappa ($\kappa$) as a measure of inter-annotator agreement to assess the reliability of our annotations. We calculate $\kappa$ over the remaining annotated test set, resulting in a $\kappa$ value of 0.858, indicating substantial agreement.

\definecolor{Mycolor1}{HTML}{BAD8F2}
\definecolor{Mycolor2}{HTML}{DDEEFA}

\begin{table*}[!hpt]
    \center 
    \small
    \scalebox{0.78}{
    \begin{tabular}
    {ll|cccccccc|cc}
        \hline
        \toprule
            & \multirow{2}{*}{\textbf{Evaluator}}  
            & \multicolumn{2}{c}{\textsc{Vanilla}} & \multicolumn{2}{c}{\textsc{Multi-hops}} 
            & \multicolumn{2}{c}{\textsc{Comparison}} 
            & \multicolumn{2}{c|}{\textsc{Set-Operation}}
            & \multicolumn{2}{c}{\textsc{Average}} \\
            \cmidrule(lr){3-4} \cmidrule(lr){5-6} \cmidrule(lr){7-8}\cmidrule(lr){9-10}
            \cmidrule(lr){11-12}
            &  
            & \textbf{\textsc{cls.}}  & \textbf{\textsc{exp.}}
            & \textbf{\textsc{cls.}}  & \textbf{\textsc{exp.}}
            & \textbf{\textsc{cls.}}  & \textbf{\textsc{exp.}}
            & \textbf{\textsc{cls.}}  & \textbf{\textsc{exp.}}
            & \textbf{\textsc{cls.}}  & \textbf{\textsc{exp.}}
            \\
     
        \midrule
            \parbox[t]{2mm}{\multirow{5}{*}{\rotatebox[origin=c]{90}{ Zero-Shot}}}
            & GPT-3.5-turbo
            & 55.12 & 22.79
            & \colorbox{Mycolor1}{59.54} & 29.84 
            & 16.66 & 18.89 
            & \colorbox{Mycolor2}{55.46}& 28.23 
            & 52.82 & 24.03   \\ 
            
            &  text-davinci-003
            & \colorbox{Mycolor2}{52.06} & 17.72
            & \colorbox{Mycolor1}{59.92} & 25.30
            & 25.50 & 16.09 
            & 48.58 & 25.71 
            & 50.98 & 19.57 
            \\
            & Alpaca-7B    
            & \colorbox{Mycolor1}{29.66} & 11.72
            & 5.20 & 25.60
            & 8.88 & 17.95 
            & \colorbox{Mycolor2}{13.08} & 21.37
            & 23.10 & 13.66 \\
            
            & Vicuna-7B  
            % & \colorbox{Mycolor1}{32.28}  & 30.46 
            % & \colorbox{Mycolor2}{17.54}   & 34.60 
            % & 9.34    & 25.79 
            % & 14.95   & 34.96 
            % & 28.86   & 31.10  \\
            & \colorbox{Mycolor1}{35.26} & 24.62
            & \colorbox{Mycolor2}{17.54} & 34.39 
            & 9.34 & 24.88 
            & 14.96 & 31.41 
            
            & 28.84 & 26.84 \\
            % & Falcon-7B   
            % & \colorbox{Mycolor1}{28.28} & 18.95 
            % & 17.46 & 26.33 
            % & 4.38 & 16.38 
            % & \colorbox{Mycolor2}{25.14} & 29.95 
            % & 24.14 & 21.01\\
            % & Alpaca-13B    
            % & 64.42   & 11.86  
            % & 64.96   & 8.68  
            % & 68.80   & 9.08
            % & 65.58   & 7.83 
            % & 65.04   & 10.66 \\
            & Llama2-7B-chat    
            &  3.57  &  26.78
            & 5.49   & 33.87  
            & \colorbox{Mycolor1}{10.53}   &  35.25  
            & \colorbox{Mycolor1}{12.61}   &   33.27 
            & 5.77   &  29.41
            \\
        \midrule
            \parbox[t]{2mm}{\multirow{5}{*}{\rotatebox[origin=c]{90}{ ICL (4-shot)}}} 
            % & GPT-3.5-turbo 
            %       & 60.04 & 45.05 & 62.62 & 53.14  & 47.18 & 44.16 & 62.34 & 53.81 & 59.64 & 47.13 \\
            & GPT-3.5-turbo 
            & 62.02 & 37.29
            & \colorbox{Mycolor1}{65.66} & 51.85 
            & 32.2 & 48.11 
            & \colorbox{Mycolor2}{64.74} & 50.14 
            & \ua{8.22}61.04 & \ua{17.93}41.96
            \\
            & text-davinci-003
            & \colorbox{Mycolor1}{56.52} & 39.36 
            & \colorbox{Mycolor2}{55.02} & 58.22 
            & 8.50 & 48.53 
            & 50.34 & 51.82 
            & \ua{1.9}52.88 & \ua{24.88}44.45 \\
            & Alpaca-7B  
            & \colorbox{Mycolor1}{35.82} & 31.01
            & \colorbox{Mycolor2}{18.12} & 40.16 
            & 8.86 & 29.28
            & 6.70 & 31.52 
            & \ua{5.24}28.34 & \ua{16.76}32.23 \\
           
           & Vicuna-7B    
            & \colorbox{Mycolor1}{41.36} & 42.51 
            & \colorbox{Mycolor2}{29.24} & 58.35 
            & 19.36 & 41.55
            & 13.46 & 53.60
            & \ua{6.3}35.14 & \ua{19.14}45.98 \\
           % & Alpaca-13B   
           %  & 67.72   & 31.42 
           %  & 63.90   & 38.69 
           %  & 51.26   & 30.17 
           %  & 57.82   & 42.70
           %  & 64.70   & 33.64 \\ 
                   & Llama2-7B-chat    
            & \colorbox{Mycolor2}{31.00}  &  39.08
            & \colorbox{Mycolor1}{39.13}  & 54.38 
            & 10.50    &  41.83
            &  {27.96} &  51.73
             & \ua{24.48}30.25 & \ua{13.61}43.02 \\
        \midrule
            \parbox[t]{2mm}{\multirow{3}{*}{\rotatebox[origin=c]{90}{Det.(tune)}}} 
            & Alpaca-7B-LoRA    
            % & \colorbox{Mycolor1}{73.16}   & 54.88  
            % & 63.34   & 73.02 
            % & \colorbox{Mycolor2}{69.92}   & 55.38
            % & 68.18   & 65.0 
            % & \ua{47.54}70.66   & 58.72 \\
            & \colorbox{Mycolor1}{73.14} & 49.00 
            & 63.34 & 70.83 
            & \colorbox{Mycolor2}{69.92} & 59.88 
            & 68.18 & 63.75 
            & \ua{42.32}70.66 & \ua{22.73}54.96 \\
            
           & Vicuna-7B-LoRA    
            & \colorbox{Mycolor1}{73.52} & 48.07
            & 64.72 & 71.74
            & \colorbox{Mycolor2}{67.34} & 62.08
            & 50.36 & 66.04 
            & \ua{34.44}69.58 & \ua{9.00}54.98 \\
           % & Falcon-7B-LoRA    
           %  & \colorbox{Mycolor1}{65.34} & 45.95 
           %  & \colorbox{Mycolor2}{36.56} & 68.88 
           %  & 28.74 & 55.80
           %  & 27.30 & 61.65 
           %  & \ua{32.16}56.30 & \ua{31.06} 52.07 \\
          % & Alpaca-13B-LoRA    
          %   & 74.28   & 55.21  
          %   & 66.24   & 72.84  
          %   & 69.48   & 55.83
          %   & 67.14   & 66.73 
          %   & 71.64   & 59.14 \\

            %         & Llama2-7B-chat-LoRA
            % & 77.28   & 49.71  
            % & 86.40   & 78.69  
            % &  80.88  & 65.52
            % & 86.32   & 71.23  
            % & \ua{76.81}79.97   & \ua{47.63}57.98 \\
            & Llama2-7B-chat-LoRA
            & \colorbox{Mycolor2}{77.41}   & 47.91  
            & 67.70   & 67.30  
            &  62.27  & 57.03
            & \colorbox{Mycolor1}{78.68}   & 65.94  
            & \ua{44.48}74.73   & \ua{10.69}53.71 \\

        \midrule
            \parbox[t]{2mm}{\multirow{5}{*}{\rotatebox[origin=c]{90}{Knowledge}}} 
            & Alpaca-7B-LoRA (wiki)
           & \colorbox{Mycolor1}{73.86} & 49.44 
           & 67.3 & 69.97
           & \colorbox{Mycolor2}{68.24} & 60.25 
           & 67.38 & 63.00
           & \ua{0.66}71.32 & \ua{0.11}55.07 \\
           
           & Vicuna-7B-LoRA  (wiki)
            % & \colorbox{Mycolor2}{75.16}   & 54.03 
            % & \colorbox{Mycolor2}{65.46}   & 73.20 
            % & 65.10   & 60.38 
            % & 55.50   & 67.04
            % & \ua{42.02}70.88   & 58.95  \\ 
            & \colorbox{Mycolor1}{75.14} & 49.56
            & \colorbox{Mycolor2}{65.46} & 72.71 
            & 65.10 & 63.51 
            & 55.42 & 66.65 
            & \ua{1.28} 70.86 & \ua{1.30}56.28 \\
          %  & Falcon-7B-LoRA   
          % & 36.58 & 29.95 
          % & 12.54 & 41.13 
          % & 11.22 & 39.42 
          % & 4.18 & 30.51 
          % & 27.92 & 32.6 \\
          %  % &  &  &  &  & &  & & & &   \\ 
            & Llama2-7B-chat-LoRA (wiki)
            & \colorbox{Mycolor2}{77.14}   & 46.71 
            &  {69.61}  & 64.17  
            &  66.05  & 49.73
            & \colorbox{Mycolor1}{78.08}  & 64.52 
            % & \ua{xx} 75.16  & \ua{xx}51.58\\
            & \ua{1.13} 75.86  & \ua{0.87}54.58\\
\cmidrule{2-12}

            % & GPT-3.5-turbo (wiki) 
            % & 55.12 & 22.79
            % & \colorbox{Mycolor1}{59.54} & 29.84 
            % & 16.66 & 18.89 
            % & \colorbox{Mycolor2}{55.46}& 28.23 
            % & 52.82 & 24.03   \\ 
            % & GPT-3.5-turbo (wiki) 
            % &  {58.34}  & 23.56
            % &  \colorbox{Mycolor1}{62.35}  & 31.25
            % &  24.14  & 21.45
            % &  \colorbox{Mycolor2}{59.37}  & 32.58
            % &  \ua{7.59}68.63  & \ua{0.41}42.37 \\
            
            & GPT-3.5-turbo (tool) 
            &  \colorbox{Mycolor2}{69.71}  & 38.60
            &  \colorbox{Mycolor1}{69.92}  & 48.43
            &  44.08  & 47.26
            &  74.21  & 45.65
            &  \ua{7.59}68.63  & \ua{0.41}42.37 \\
                   \midrule
            \GG \parbox[t]{2mm}{\multirow{1}{*}{\rotatebox[origin=c]{90}{}}} 
            & \GG \textsc{\ourmodel} 
           & \GG 80.97  &\GG 47.08
           &\GG 75.01  &\GG 64.21
           & \GG 66.27  &\GG  55.70
           & \GG 80.87 &\GG 65.25
           & \GG 78.15 &\GG  52.52 \\ 
        \bottomrule
        \hline
    \end{tabular}
    }
    \caption{
      \small
        Results on \textsc{FactCls} and \textsc{ExpMatch}     (abbreviated as  \textbf{CLS.} and \textbf{EXP.}) along with \ours{} estimated by each method. 
        The \colorbox{Mycolor1}{shadow} and \colorbox{Mycolor2}{shadow} in each row represent the top-2 \textsc{FactCls} scores for the four factuality patterns. 
        The \protect\ua{up} and \protect\da{down} arrows respectively indicate positive/negative performance changes in the \textsc{Average} score compared to the corresponding upper-level method.
        % We adopt consistent prompts in all evaluation experiments for a fair comparison.
    }
    \label{tab:main_results}
\end{table*}

\section{Experiments}

\subsection{Metric Definition}

\paragraph{\textbf{\textsc{FactCls} Metric}.}
We employ the \textsc{FactCls}, denoted by the Micro F1 score, to evaluate binary factuality classification performance. This metric focuses on the distribution $p(l|Q\&R)$, classifying instances as either \textsc{factual} or \textsc{non-factual}. With a specific emphasis on identifying non-factual examples, we designate \textsc{non-factual} as the positive class and \textsc{factual} as the negative class.

\paragraph{\textbf{\textsc{ExpMatch} Metric}.}
In {\ours}, the golden evidence chain features introductory/expository statements (head-tail part) and a factual explanation core (body part). The former contextualizes with phrases such as `Therefore, there is an incorrect conclusion in this query and response', and the latter delivers the in-depth, fact-based reasoning process. For hallucination detectors, prompts guide their outputs akin to the gold evidence chain for quality assessment.
Given the paramount importance of aligning factual explanations over expository parts, we introduce \textsc{ExpMatch}, a metric employing segmented matching with weighted averaging. It computes $\text{Score}_{bd}$ via span-based Micro F1 for unigram overlap between generated and reference bodies, and $\text{Score}_{ht}$ via ROUGE-L to assess similarity based on the longest common word subsequence for the head-to-tail part. \textsc{ExpMatch}, combining $\text{Score}_{bd}$ and $\text{Score}_{ht}$,  evaluates the explanations generated from detectors as:
\begin{equation}
\small
\textsc{ExpMatch} = \alpha \times \text{Score}_{bd} + (1-\alpha) \times \text{Score}_{ht}.
\end{equation}
We initially set $\alpha$ to 0.7, emphasizing the body part\footnote{Preliminary evaluations reveal that scores with $\alpha$ at 0.75 and 0.8 align with our established conclusions. If a generated result lacks factual or non-factual information, indicating a prediction failure, we attribute an \textsc{ExpMatch} value of 0 to that particular example during metric calculation.}.

\subsection{Experimental Settings}

\paragraph{Evaluation Models.}

We evaluate various leading LLMs on {\ours} benchmark, focusing on OpenAI API models, including \texttt{text-davinci-003} (InstructGPT) and \texttt{GPT-3.5-turbo} (ChatGPT). Additionally, we explore the adoption of open-source models such as Llama2-chat\footnote{\url{https://huggingface.co/meta-llama/Llama-2-7b-chat}}, Alpaca\cite{alpaca} and Vicuna \cite{vicuna2023}, which are fine-tuned variants of the LLaMA \shortcite{llama2023}.

\begin{figure*}[htb] %H为当前位置，!htb为忽略美学标准，htbp为浮动图形
\centering %图片居中
\includegraphics[width=0.9\textwidth]{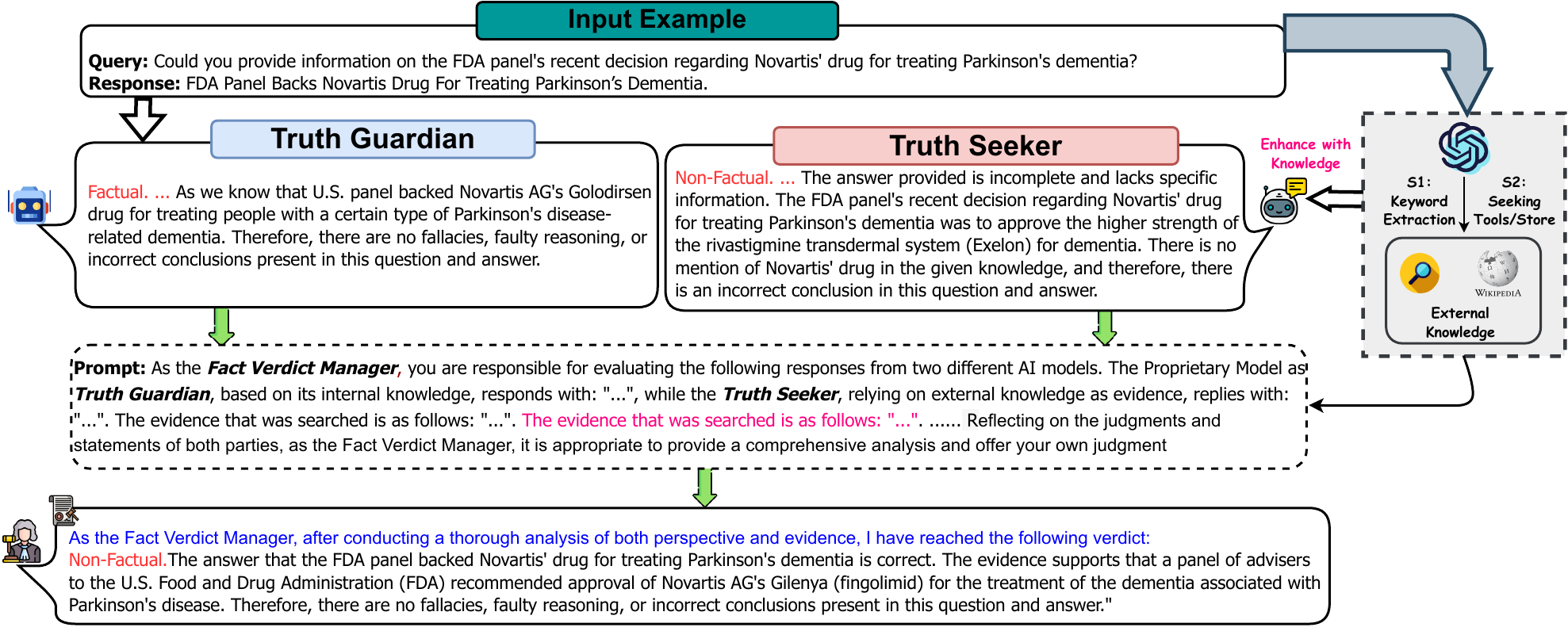} %
% 插入图片，[]中设置图片大小，{}中是图片文件名
  \caption{\small  Overview \ourmodel. Here we designate the 
    {\emojitruth} as the ``Truth Guardian'' based on Llama2-7B-chat-LoRA while {\emojiseeker} as the ``Truth Seeker'' based on GPT-3.5-turbo (tool) in our experiments.
     We want the {\emojijudge}  ``Fact Verdict Manager'' to collect evidence from different viewpoints to enhance the reliability and accuracy of the obtained conclusion.
}
\label{fig:truth}
\end{figure*}

\paragraph{Implementation Details.}
Using Azure's OpenAI ChatGPT API, we generate samples with a temperature of $1.0$ to control the diversity of generated samples, while limiting the maximum number of tokens to $2048$ to ensure concise responses.
We use a frequency penalty of zero and a Top-$p$ of $1.0$ to ensure unrestricted token selection during generation. For evaluations, we standardize the temperature at $0.2$ to minimize randomness. 
% Detailed instructions are provided in Table \ref{tab:prompt1} through Table \ref{tab:prompt2} of \S\ref{app:exp_instruct}.

\paragraph{Baseline Strategy Settings}
This study investigates the effectiveness of various baseline strategies in detecting fact-conflicting hallucinations. These strategies include:
\textsc{(1) Zero-shot Learning}, which evaluates model performance without prior training in hallucination detection;
\textsc{(2) In-context Learning}, using 4-shot samples as demonstrations to prompt the model and evaluate its ability to handle hallucination detection;
\textsc{(3) Detect-Specific Expert Model}, fine-tuning the LoRA~\cite{lora} parameters using our specialized training set to leverage domain expertise;
\textsc{(4) Knowledge Enhancement}, enhancing the model's capabilities by integrating external knowledge through retrieval techniques or tool-based enhancements.
These experiments aim to explore both the potential and limitations of these baseline methods in hallucination detection.

\subsection{Empirical Experiment Results}

\subsubsection{Zero-shot Learning Performance}
The empirical results presented in Table~\ref{tab:main_results} show that LLMs with zero-shot learning struggle to identify implicit factuality in ``\textsc{Query-Response}'' contexts. The performance of the open-source 7B LLMs in zero-shot learning is notably poor, indicating deficiencies in both instruction comprehension and internal knowledge representation. Even the ChatGPT shows limited proficiency in distinguishing between factual and non-factual ``\textsc{Query-Response}'' samples, achieving only a 52.82\% FactCls score in zero-shot.  Llama2-chat is susceptible to false negatives, resulting in reduced sensitivity in hallucination detection.

\subsubsection{In-context Learning Performance}

The incorporation of few-shot information significantly improves fact-conflicting hallucination detection in the GPT-3.5-turbo, Alpaca-7B, Vicuna-7B, and Llama2-7B-chat models. Compared to models operating without this additional context, these enhancements result in an average increase of approximately \daulg{6\%} in the \textsc{FactCls}  and \daulg{18\%} in \textsc{ExpMatch} scores. However, the impact of integrating few-shot information on \textit{text-davinci-003} is relatively modest, indicating its limited proficiency in managing few-shot learning. The improvement from incorporating few-shot information into Llama2-chat is significant, possibly because of the model's superior contextual learning abilities compared to Alpaca and Vicuna, leading to a better understanding of  demonstrations.

\subsubsection{Detect-Specific Expert Performance} 

We investigate the effectiveness of tuning LLMs with LoRA for domain-specific expertise in detecting fact-conflicting hallucinations. By training these models on our trainset, we enhance their detection task proficiency. Our findings show that all tested open-source 7B models improve, with Llama2-chat outperforming its counterparts. After fine-tuning on hallucination data, Llama2-chat-7B achieves a \textsc{FactCls} score of 74.73\% and an \textsc{ExpMatch} score of 53.71\% on our benchmark. This success with a 7B model underscores the viability of using such models as hallucination detectors and encourages further exploration in this area.

\subsubsection{Knowledge Enhancemant}
\paragraph{Retrieval Enhancement.}

This study enhances LLM-based detection with Wikipedia-sourced facts, using BM25 for initial document retrieval to select the top five most relevant paragraphs. Our experiments show that given the 7B model's limited ability to process long texts, providing knowledge during fine-tuning modestly improves performance.  However, compared to the detect-specific expert without knowledge augmentation, the improvement achieved by combining Wikipedia retrieval with fine-tuning the LoRA parameters is relatively modest. We attribute these observations to two primary factors: our dataset spans multiple domains while Wikipedia encompasses merely a subset, and our retrieval method being elementary, occasionally yields lower-quality evidence.

\begin{table*}[!]
    \centering
    % \small
    \resizebox{1.0\linewidth}{!}{
    \begin{tabularx}{1.5\linewidth}{bbb}
         \toprule[1.5pt] \textbf{Query and Returned Evidence} 
          & \textbf{Detection Opinion from Both Sides}
         & \textbf{Prediction with Triangulation} \\ 
         \midrule[0.75pt]
          \textbf{\textsc{Query: }} Who was the mother of Afonso II, the third king of Portugal? 
          
     \textbf{\textsc{Response:}}The mother of Afonso II, the third king of Portugal, was  \textcolor{red!90!white}{ Queen Urraca of Castile.}
         
         \textbf{Correct Answer:} \textbf{Dulce Berenguer of Barcelona.}
         
          \textbf{Returned by the tool:} \textcolor{gold}{
           1.—died March 25, 1223, Coimbra), the third king of Portugal (1211–23), under whom the reconquest of the south from the Muslims was continued. born: 1185?, Coimbra, Port. Afonso II was the son of King Sancho I and Queen Dulcia, daughter of Ramón Berenguer IV of Barcelona. 2...}

         & \textbf{\textsc{Truth Guardian:}}
         \textcolor{red!90!white}{FACTUAL.} 
        The answer that Queen Urraca of Castile was the mother of Afonso II, the third king of Portugal, is correct. This is supported by the evidence that Afonso II was the third king of Portugal and that Queen Urraca of Castile was his mother. Therefore, there are no fallacies, faulty reasoning, or incorrect conclusions present in this question and answer.
        
         \textbf{\textsc{Truth Seeker:}}
         \textcolor{green!90!white}{NON-FACTUAL.} 
         The answer stating that the mother of Afonso II, the third king of Portugal, was Queen Urraca of Castile is incorrect. According to the provided knowledge, the mother of Afonso II was Queen Dulcia, daughter of Ramón Berenguer IV of Barcelona. Therefore, there is an incorrect conclusion in this question and answer.

         &\textbf{\ourmodel:}
           As the Fact Verdict Manager, after conducting a thorough analysis of both perspectives and evidence, I have reached the following verdict:
         \textcolor{green!90!white}{NON-FACTUAL.} The answer that Queen Urraca of Castile was the mother of Afonso II, the third king of Portugal, is incorrect. The evidence provided shows that Afonso II was the son of King Sancho I and Queen Dulcia, daughter of Ramón Berenguer IV of Barcelona. Therefore, the answer contains false assumptions and incorrect conclusions. The Truth Seeker's response is supported by the provided evidence, while the Proprietary Model's response is not accurate as it did not consider the external evidence. \\

   \midrule[0.75pt]
         \textbf{\textsc{Query:}} Who is the Jeff Probst Show-nominated television producer's first wife?

         \textbf{\textsc{Response:}}His first wife was psychotherapist \textcolor{blue!90!white}{Shelley Wright.}  
         
          \textbf{Correct Answer:} \textbf{Shelley Wright.}

         \textbf{Returned by the tool:}\textcolor{gold}
         {
           1. Probst was married to his first wife, psychotherapist Shelley Wright, from 1996 to 2001. In 2004, after filming on Survivor: Vanuatu wrapped, he began dating one of the contestants, a 23 year old Julie Berry - he was about 20 years her senior. They broke up in early 2008. 2. ...}

         &  \textbf{\textsc{Truth Guardian:}
         \textcolor{green!90!white}}{FACTUAL.}
         The answer that Jeff Probst's first wife was psychotherapist Shelley Wright is correct. According to the evidence provided, Jeff Probst was previously married to Shelley Wright, who is a *psychotherapist* in addition to being a television producer. Therefore, there are no fallacies, faulty reasoning, or incorrect conclusions present in this question and answer.

         \textbf{\textsc{Truth Seeker:}}
         \textcolor{green!90!white}{FACTUAL.} 
          According to the provided knowledge, Jeff Probst's first wife was psychotherapist Shelley Wright. Therefore, the answer is correct and there are no fallacies, faulty reasoning, or incorrect conclusions present in this question and answer.
         
     &\textbf{\ourmodel:}
        As the Fact Verdict Manager, after evaluating the responses from both AI models and examining the provided evidence, I have reached the following verdict:

 \textcolor{green!90!white}{FACTUAL.}  The answer that Jeff Probst's first wife was psychotherapist Shelley Wright is correct. Both AI models correctly identified that there are no fallacies, faulty reasoning, or incorrect conclusions present in the question and answer. The evidence provided confirms that Jeff Probst was married to Shelley Wright from 1996 to 2001, and that she is a psychotherapist in addition to being a television producer. Therefore, the answer is factual and accurate. \\
    \bottomrule[1.5pt]    
    \end{tabularx}
    }
 \caption{Case analysis of out-of-distribution examples from ChatGPT using {\ourmodel}.}
    \label{tab:main_case}
\end{table*}

\paragraph{Tool Enhancement.}
% Considering the labor-intensive nature of tapping into external knowledge bases, we extend our exploration into exploiting ChatGPT's potent contextual capabilities for tool-enhanced learning. Specifically inspired by preceding works such as~\cite{factool,too-foundation}, we harness the tool's execution results to bolster large-scale models in hallucination detection. Employing a search engine as a tool to address queries from the ``\textsc{Query-Response}'' samples, we prompt ChatGPT to ascertain hallucination presence in the input ``\textsc{Query-Response}'' grounded on the search engine's retrieved results, thereby providing concomitant justifications. Experimental outcomes, detailed in Table~\ref{tab:main_results}, not only showcase this method’s substantial improvement over the zero-shot approach predicated on ChatGPT but also further underscore the promising potential imbued in tool-enhanced learning.

Considering the labor-intensive nature of tapping into external knowledge bases,
we investigate leveraging ChatGPT's advanced contextual understanding for tool-enhanced hallucination detection, drawing from prior research such as~\cite{factool,too-foundation}. By using a search engine to inform ChatGPT's analysis of ``\textsc{Query-Response}'' samples, we enable the model to identify hallucinations with supporting evidence from search results. Results in Table~\ref{tab:main_results} demonstrate significant enhancements over ChatGPT's few-shot performance, highlighting the efficacy of tool learning.

\begin{figure}[tb]
    \centering
    \includegraphics[width=0.67\linewidth]{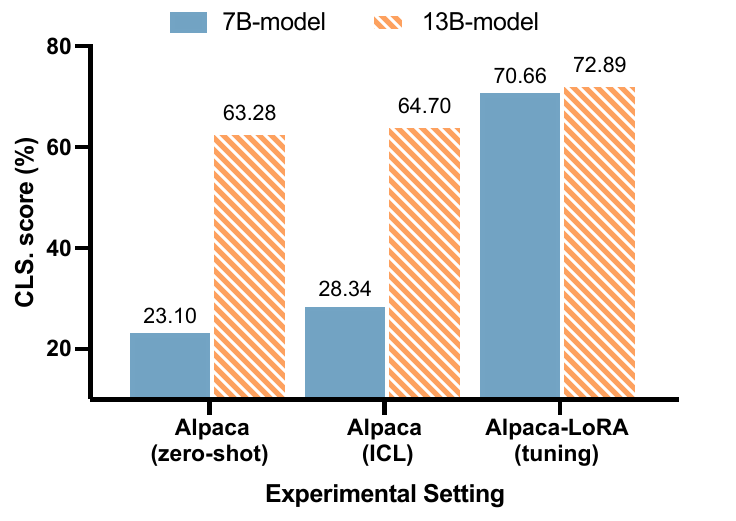}
    \caption{ Analysis of model capacity impact. }
    \label{fig:model_cap}
\end{figure}

\subsection{Exploring Triangulation for Truth}
We categorize tool-enhanced ChatGPT as the \textbf{Truth Seeker}, which aims to make informed judgments by seeking external knowledge. However, the information returned by external knowledge sources may inevitably be incomplete, erroneous, or redundant, thus potentially misleading the large-scale model. On the other hand, the detect-specific expert as the \textbf{Truth Guardian} relies on its knowledge and expertise in the task, tending towards more conservative predictions. To address these challenges, we propose the {\ourmodel} framework inspired by the ``Triangulation for Truth'' theory, involving verifying and confirming information by cross-referencing multiple independent perspectives. 
Figure~\ref{fig:truth} illustrates our approach of fine-tuning ChatGPT as a \textbf{Fact Verdict Manager}, leveraging the output and knowledge from \textbf{Truth Seeker} and \textbf{Truth Guardian} to boost conclusion reliability. 
Table \ref{tab:main_results} reveals our model's superior performance to Lamma2-7b-chat-LoRA and GPT-3.5-turbo (tool), highlighting the benefits of triangulation in reducing single-source error and enhancing truth verification.

% As shown in Figure~\ref{fig:truth}, we fine-tune ChatGPT with over 10 examples as a \textbf{Fact Verdict Manager} and collect evidence from \textbf{Truth Seeker} and \textbf{Truth Guardian} to enhance the reliability and accuracy of the obtained conclusion. Table \ref{tab:main_results} demonstrates that our model, compared to Lamma2-7b-chat-LoRA and GPT-3.5-turbo (tool), exhibits improvements. This finding further underscores the effectiveness of triangulation in mitigating errors and inconsistencies that may arise from relying on a single source or method, thereby facilitating a more comprehensive and robust understanding of the truth.

% \begin{equation}
%     J_d(H)= \begin{cases}
%         \texttt{True},~~ & \textrm{solution obtained} \\
%         \texttt{False},~~ & \textrm{otherwise}
%     \end{cases}.
% \end{equation}

\subsection{Experimental Analysis}

\paragraph{Examining the Influences of Model Capacity.}

Figure~\ref{fig:model_cap} illustrates that transitioning from 7B to 13B models notably improves the detection of fact-conflicting hallucinations, particularly in zero-shot and in-context learning scenarios. Alpaca-13B outperforms ChatGPT, which can beascribed to the consistently adopted command prompt may be more friendly to Alpaca-13B~\footnote{We keep consistent prompts for each LLM throughout all experiments. Given that LLMs are acknowledged for their pronounced sensitivity to prompts, we do not assert that our prompts are universally optimal.}. 
Interestingly, when models are fine-tuned with training data, the impact of model capacity on performance improvement appears minimal. This implies that further training larger LLMs as hallucination detectors may yield limited benefits,
and it is necessary to explore alternatives with higher upper limits for enhancing detection performance.

\paragraph{Enhancement lies in Accurate Evidence.}

We also explore the direct use of intrinsic facts from the dataset as golden evidence in the LoRA fine-tuning process. As shown in Table ~\ref{tab:analysis}, integrating factual information (\textbf{w/ gold evidence}) leads to a significant improvement in the \textsc{FactCls} score, indicating that the modest improvement from retrieval may stem from the lower quality of the obtained evidence. This underscores the potential for substantial improvement by seeking the most accurate facts for evaluation.

% A distinct characteristic of our dataset, which sets it apart from traditional fact-checking datasets, involves incorporating the ``\textsc{Query-Response}'' context. To delve deeper into its significance, we execute experiments, omitting the ``{query}'' during the Alpaca-7B-LoRA fine-tuning process.
% Table~\ref{tab:analysis} displays the outcomes, wherein the ``\textbf{w/o query}'' scenario leads to a substantial decline of approximately 40\% in the \textsc{FactCls} score, underscoring the pivotal role of a comprehensive ``\textsc{Query-Response}'' context in extracting invaluable information for informed decision-making.

\paragraph{Complete ``\textsc{Query-Response}'' Context.}

Our dataset differs from traditional fact-checking datasets in its inclusion of the ``\textsc{Query-Response}'' context. To examine its impact, we conduct experiments by excluding the ``query'' during the Alpaca-7B-LoRA fine-tuning process. The results in Table~\ref{tab:analysis} reveal that the ``\textbf{w/o query}'' scenario leads to a significant 40\% decrease in the \textsc{FactCls} score, highlighting the essential role of a comprehensive ``\textsc{Query-Response}'' context in extracting valuable information for informed decision-making.

\begin{table}[t]
	\small
	\centering

 \scalebox{0.82}{
	\setlength\tabcolsep{3.5pt}
	\begin{tabular}{c| c c c c|c}
		\toprule
		  \textbf{Variants} & \textsc{Conv.} & \textsc{Multi.} & \textsc{Comp.} & \textsc{Set.} & \textsc{Avg.} \\
	\midrule
		{Alpaca-7B-LoRa} 
                & 73.16  
                & 63.34 
                & 69.92 
                & 68.18 
                & 70.66    \\
            \midrule
		    w/ golden evidence 
                & 82.68
                & 96.10
                & 70.50 
                & 87.90 
                & 83.56   \\ 
            % w/o chains 
            %     & 74.50
            %     & 79.12
            %     & 74.42 
            %     & 79.54 
            %     & 75.78   \\ 
             w/o query  
                & 39.0 
                & 10.04 
                & 9.24 
                & 9.52
                & 30.48   \\
		\bottomrule
	\end{tabular}}
        % \vspace{-0.3cm}
         \caption{ Ablation analysis on input context.}
	\label{tab:analysis}
\end{table}

% \paragraph{Case Analysis.}
% To underscore the broad applicability of {\ourmodel}, we embarked on supplementary tests, applying it to \emph{real-world hallucination data from ChatGPT} that transcends the boundaries of the {\ours} benchmark. We elucidate both the capabilities and constraints of our model by showcasing the findings from our out-of-distribution case analysis in Table \ref{tab:main_case}. The results from these cases substantiate that our method can make proficient judgments, particularly when discrepancies arise between the detect-specific expert and tool-enhanced ChatGPT. This enhances the reliability of fact-conflicting hallucination detection by providing a practical examination of its application in genuine, unscripted scenarios.

\paragraph{Real-World Application Case Study.}
We extended the testing of {\ourmodel} to real-world instances of ChatGPT-generated hallucinations beyond the {\ours} benchmark to demonstrate its wide-ranging utility. The out-of-distribution case analysis, detailed in Table \ref{tab:main_case}, delineates our model's strengths and limitations. These real-world tests confirm {\ourmodel}'s adeptness in making accurate assessments, particularly where expert and augmented ChatGPT evaluations diverge, thus bolstering the credibility of detecting fact-conflicting hallucinations in authentic, uncontrolled settings.

% This significantly improves the reliability of fact-hallucination detection by testing its application in real-world scenarios.

\section{Conclusion and Future Work}
We introduce {\ours}, a meticulously designed benchmark tailored for the evaluation of fact-conflicting hallucinations from LLMs, which is notably enriched with a kaleidoscope of patterns and substantiated evidence chains to fortify the robust elucidation of factuality assessments.
Moreover, we delineate {\ourmodel} that employs the principle of triangulation to discern the veracity of information, deploying cross-referencing generators to arbitrate responses. Moving forward, 
% our research will concentrate on exploring efficient knowledge-enhancement methods to enhance hallucination detection.
we will broaden our evaluative scope to encompass various modalities and granularity in hallucination detection.

\section*{Acknowledgments}
We would like to express gratitude to the anonymous reviewers for kind comments. 
This work was supported by the National Natural Science Foundation of China (No. 62206246), Response-driven intelligent enhanced control technology for AC/DC hybrid power grid with high proportion of new energy (5100-202155426A-0-0-00), the Fundamental Research Funds for the Central Universities (226-2023-00138), Zhejiang Provincial Natural Science Foundation of China (No. LGG22F030011), Yongjiang Talent Introduction Programme (2021A-156-G), CCF-Tencent Rhino-Bird Open Research Fund, and Information Technology Center and State Key Lab of CAD\&CG, Zhejiang University.

% \clearpage
\bibliographystyle{named}
\bibliography{ijcai24}

\begin{thebibliography}{}

\bibitem[\protect\citeauthoryear{Aly \bgroup \em et al.\egroup }{2021}]{FEVEROUS}
Rami Aly, Zhijiang Guo, Michael~Sejr Schlichtkrull, James Thorne, Andreas Vlachos, Christos Christodoulopoulos, Oana Cocarascu, and Arpit Mittal.
\newblock {FEVEROUS:} fact extraction and verification over unstructured and structured information.
\newblock In Joaquin Vanschoren and Sai{-}Kit Yeung, editors, {\em Proceedings of the Neural Information Processing Systems Track on Datasets and Benchmarks 1, NeurIPS Datasets and Benchmarks 2021, December 2021, virtual}, 2021.

\bibitem[\protect\citeauthoryear{Chandak \bgroup \em et al.\egroup }{2023}]{chandak2023building}
Payal Chandak, Kexin Huang, and Marinka Zitnik.
\newblock Building a knowledge graph to enable precision medicine.
\newblock {\em Scientific Data}, 10(1):67, 2023.

\bibitem[\protect\citeauthoryear{Chen \bgroup \em et al.\egroup }{2020}]{TabFact}
Wenhu Chen, Hongmin Wang, Jianshu Chen, Yunkai Zhang, Hong Wang, Shiyang Li, Xiyou Zhou, and William~Yang Wang.
\newblock Tabfact: {A} large-scale dataset for table-based fact verification.
\newblock In {\em {ICLR} 2020}. OpenReview.net, 2020.

\bibitem[\protect\citeauthoryear{Chen \bgroup \em et al.\egroup }{2022}]{knowprompt}
Xiang Chen, Ningyu Zhang, Xin Xie, Shumin Deng, Yunzhi Yao, Chuanqi Tan, Fei Huang, Luo Si, and Huajun Chen.
\newblock Knowprompt: Knowledge-aware prompt-tuning with synergistic optimization for relation extraction.
\newblock In {\em {WWW} 2022}, pages 2778--2788. {ACM}, 2022.

\bibitem[\protect\citeauthoryear{Chern \bgroup \em et al.\egroup }{2023}]{factool}
I{-}Chun Chern, Steffi Chern, Shiqi Chen, Weizhe Yuan, Kehua Feng, Chunting Zhou, Junxian He, Graham Neubig, and Pengfei Liu.
\newblock Factool: Factuality detection in generative {AI} - {A} tool augmented framework for multi-task and multi-domain scenarios.
\newblock {\em CoRR}, abs/2307.13528, 2023.

\bibitem[\protect\citeauthoryear{Chiang \bgroup \em et al.\egroup }{2023}]{vicuna2023}
Wei-Lin Chiang, Zhuohan Li, Zi~Lin, Ying Sheng, Zhanghao Wu, Hao Zhang, Lianmin Zheng, Siyuan Zhuang, Yonghao Zhuang, Joseph~E. Gonzalez, Ion Stoica, and Eric~P. Xing.
\newblock Vicuna: An open-source chatbot impressing gpt-4 with 90\%* chatgpt quality, March 2023.

\bibitem[\protect\citeauthoryear{Creswell and Shanahan}{2022}]{creswell2022faithful}
Antonia Creswell and Murray Shanahan.
\newblock Faithful reasoning using large language models.
\newblock {\em arXiv preprint arXiv:2208.14271}, 2022.

\bibitem[\protect\citeauthoryear{Das \bgroup \em et al.\egroup }{2023}]{dialogue-hallucination}
Souvik Das, Sougata Saha, and Rohini~K Srihari.
\newblock Diving deep into modes of fact hallucinations in dialogue systems.
\newblock {\em arXiv preprint arXiv:2301.04449}, 2023.

\bibitem[\protect\citeauthoryear{Dhingra \bgroup \em et al.\egroup }{2019}]{DhingraFPCDC19}
Bhuwan Dhingra, Manaal Faruqui, Ankur~P. Parikh, Ming{-}Wei Chang, Dipanjan Das, and William~W. Cohen.
\newblock Handling divergent reference texts when evaluating table-to-text generation.
\newblock In Anna Korhonen, David~R. Traum, and Llu{\'{\i}}s M{\`{a}}rquez, editors, {\em {ACL} 2019}, pages 4884--4895. Association for Computational Linguistics, 2019.

\bibitem[\protect\citeauthoryear{Diggelmann \bgroup \em et al.\egroup }{2020}]{diggelmann2020climate}
Thomas Diggelmann, Jordan Boyd-Graber, Jannis Bulian, Massimiliano Ciaramita, and Markus Leippold.
\newblock Climate-fever: A dataset for verification of real-world climate claims.
\newblock {\em arXiv preprint arXiv:2012.00614}, 2020.

\bibitem[\protect\citeauthoryear{Dziri \bgroup \em et al.\egroup }{2022}]{DziriKMZYPR22}
Nouha Dziri, Ehsan Kamalloo, Sivan Milton, Osmar~R. Za{\"{\i}}ane, Mo~Yu, Edoardo~Maria Ponti, and Siva Reddy.
\newblock Faithdial: {A} faithful benchmark for information-seeking dialogue.
\newblock {\em Trans. Assoc. Comput. Linguistics}, 10:1473--1490, 2022.

\bibitem[\protect\citeauthoryear{Gupta \bgroup \em et al.\egroup }{2022}]{GuptaWLX22}
Prakhar Gupta, Chien{-}Sheng Wu, Wenhao Liu, and Caiming Xiong.
\newblock Dialfact: {A} benchmark for fact-checking in dialogue.
\newblock In {\em {ACL} 2022}, pages 3785--3801, 2022.

\bibitem[\protect\citeauthoryear{Hu \bgroup \em et al.\egroup }{2022}]{lora}
Edward~J. Hu, Yelong Shen, Phillip Wallis, Zeyuan Allen{-}Zhu, Yuanzhi Li, Shean Wang, Lu~Wang, and Weizhu Chen.
\newblock Lora: Low-rank adaptation of large language models.
\newblock In {\em The Tenth International Conference on Learning Representations, {ICLR} 2022, Virtual Event, April 25-29, 2022}. OpenReview.net, 2022.

\bibitem[\protect\citeauthoryear{Huang \bgroup \em et al.\egroup }{2023}]{TRUSTGPT}
Yue Huang, Qihui Zhang, Philip~S. Yu, and Lichao Sun.
\newblock Trustgpt: A benchmark for trustworthy and responsible large language models.
\newblock {\em CoRR}, abs/2304.10513, 2023.

\bibitem[\protect\citeauthoryear{Ji \bgroup \em et al.\egroup }{2023}]{hallucination-survey}
Ziwei Ji, Nayeon Lee, Rita Frieske, Tiezheng Yu, Dan Su, Yan Xu, Etsuko Ishii, Ye~Jin Bang, Andrea Madotto, and Pascale Fung.
\newblock Survey of hallucination in natural language generation.
\newblock {\em ACM Computing Surveys}, 55(12):1--38, 2023.

\bibitem[\protect\citeauthoryear{Jiang \bgroup \em et al.\egroup }{2020}]{HoVer}
Yichen Jiang, Shikha Bordia, Zheng Zhong, Charles Dognin, Maneesh~Kumar Singh, and Mohit Bansal.
\newblock Hover: {A} dataset for many-hop fact extraction and claim verification.
\newblock In {\em Findings of {EMNLP} 2020}, 2020.

\bibitem[\protect\citeauthoryear{Kang \bgroup \em et al.\egroup }{2024}]{CRAG}
Mintong Kang, Nezihe~Merve G{\"{u}}rel, Ning Yu, Dawn Song, and Bo~Li.
\newblock {C-RAG:} certified generation risks for retrieval-augmented language models.
\newblock {\em CoRR}, abs/2402.03181, 2024.

\bibitem[\protect\citeauthoryear{Kry{\'s}ci{\'n}ski \bgroup \em et al.\egroup }{2020}]{kryscinski2020evaluating}
Wojciech Kry{\'s}ci{\'n}ski, Bryan McCann, Caiming Xiong, and Richard Socher.
\newblock Evaluating the factual consistency of abstractive text summarization.
\newblock In {\em EMNLP}, pages 9332--9346, 2020.

\bibitem[\protect\citeauthoryear{Li \bgroup \em et al.\egroup }{2023}]{HaluEval}
Junyi Li, Xiaoxue Cheng, Wayne~Xin Zhao, Jian{-}Yun Nie, and Ji{-}Rong Wen.
\newblock Halueval: {A} large-scale hallucination evaluation benchmark for large language models.
\newblock {\em CoRR}, abs/2305.11747, 2023.

\bibitem[\protect\citeauthoryear{Liang \bgroup \em et al.\egroup }{2023}]{keliang23}
Ke~Liang, Sihang Zhou, Yue Liu, Lingyuan Meng, Meng Liu, and Xinwang Liu.
\newblock Structure guided multi-modal pre-trained transformer for knowledge graph reasoning.
\newblock {\em CoRR}, abs/2307.03591, 2023.

\bibitem[\protect\citeauthoryear{Liang \bgroup \em et al.\egroup }{2024}]{tkde/LiangLZTWYDL24}
Ke~Liang, Yue Liu, Sihang Zhou, Wenxuan Tu, Yi~Wen, Xihong Yang, Xiangjun Dong, and Xinwang Liu.
\newblock Knowledge graph contrastive learning based on relation-symmetrical structure.
\newblock {\em {IEEE} Trans. Knowl. Data Eng.}, 36(1):226--238, 2024.

\bibitem[\protect\citeauthoryear{Mallen \bgroup \em et al.\egroup }{2022}]{mallen2022not}
Alex Mallen, Akari Asai, Victor Zhong, Rajarshi Das, Hannaneh Hajishirzi, and Daniel Khashabi.
\newblock When not to trust language models: Investigating effectiveness and limitations of parametric and non-parametric memories.
\newblock {\em arXiv preprint arXiv:2212.10511}, 2022.

\bibitem[\protect\citeauthoryear{Mohr \bgroup \em et al.\egroup }{2022}]{CoVERT}
Isabelle Mohr, Amelie W{\"{u}}hrl, and Roman Klinger.
\newblock Covert: {A} corpus of fact-checked biomedical {COVID-19} tweets.
\newblock In {\em {LREC} 2022}, pages 244--257, 2022.

\bibitem[\protect\citeauthoryear{Muhlgay \bgroup \em et al.\egroup }{2023}]{factor}
Dor Muhlgay, Ori Ram, Inbal Magar, Yoav Levine, Nir Ratner, Yonatan Belinkov, Omri Abend, Kevin Leyton{-}Brown, Amnon Shashua, and Yoav Shoham.
\newblock Generating benchmarks for factuality evaluation of language models.
\newblock {\em CoRR}, abs/2307.06908, 2023.

\bibitem[\protect\citeauthoryear{OpenAI}{2022}]{chatgpt}
OpenAI.
\newblock Chatgpt: Optimizing language models for dialogue.
\newblock \url{https://openai.com/blog/chatgpt}, 2022.

\bibitem[\protect\citeauthoryear{Pagnoni \bgroup \em et al.\egroup }{2021}]{pagnoni2021understanding}
Artidoro Pagnoni, Vidhisha Balachandran, and Yulia Tsvetkov.
\newblock Understanding factuality in abstractive summarization with frank: A benchmark for factuality metrics.
\newblock In {\em NAACL}, pages 4812--4829, 2021.

\bibitem[\protect\citeauthoryear{Qin \bgroup \em et al.\egroup }{2023}]{too-foundation}
Yujia Qin, Shengding Hu, and et~al.
\newblock Tool learning with foundation models.
\newblock {\em CoRR}, abs/2304.08354, 2023.

\bibitem[\protect\citeauthoryear{Rashkin \bgroup \em et al.\egroup }{2021}]{Rashkin-Measuring}
Hannah Rashkin, Vitaly Nikolaev, Matthew Lamm, Michael Collins, Dipanjan Das, Slav Petrov, Gaurav~Singh Tomar, Iulia Turc, and David Reitter.
\newblock Measuring attribution in natural language generation models.
\newblock {\em CoRR}, abs/2112.12870, 2021.

\bibitem[\protect\citeauthoryear{Reimers and Gurevych}{2019}]{sbert}
Nils Reimers and Iryna Gurevych.
\newblock Sentence-bert: Sentence embeddings using siamese bert-networks.
\newblock In {\em {EMNLP-IJCNLP} 2019}, pages 3980--3990, 2019.

\bibitem[\protect\citeauthoryear{Saakyan \bgroup \em et al.\egroup }{2021}]{covid-fact}
Arkadiy Saakyan, Tuhin Chakrabarty, and Smaranda Muresan.
\newblock Covid-fact: Fact extraction and verification of real-world claims on {COVID-19} pandemic.
\newblock In {\em {ACL/IJCNLP} 2021}, pages 2116--2129, 2021.

\bibitem[\protect\citeauthoryear{Sarrouti \bgroup \em et al.\egroup }{2021}]{healthfever}
Mourad Sarrouti, Asma~Ben Abacha, Yassine Mrabet, and Dina Demner{-}Fushman.
\newblock Evidence-based fact-checking of health-related claims.
\newblock In {\em Findings {EMNLP} 2021}, pages 3499--3512, 2021.

\bibitem[\protect\citeauthoryear{Shuster \bgroup \em et al.\egroup }{2021}]{shuster2021retrieval}
Kurt Shuster, Spencer Poff, Moya Chen, Douwe Kiela, and Jason Weston.
\newblock Retrieval augmentation reduces hallucination in conversation.
\newblock In {\em Findings of EMNLP 2021}, 2021.

\bibitem[\protect\citeauthoryear{Taori \bgroup \em et al.\egroup }{2023}]{alpaca}
Rohan Taori, Ishaan Gulrajani, Tianyi Zhang, Yann Dubois, Xuechen Li, Carlos Guestrin, Percy Liang, and Tatsunori~B. Hashimoto.
\newblock Stanford alpaca: An instruction-following llama model.
\newblock \url{https://github.com/tatsu-lab/stanford_alpaca}, 2023.

\bibitem[\protect\citeauthoryear{Thorne \bgroup \em et al.\egroup }{2018}]{thorne-etal-2018-fever}
James Thorne, Andreas Vlachos, Christos Christodoulopoulos, and Arpit Mittal.
\newblock {FEVER}: a large-scale dataset for fact extraction and {VER}ification.
\newblock In {\em NAACL (Long Papers)}, pages 809--819, New Orleans, Louisiana, June 2018.

\bibitem[\protect\citeauthoryear{Touvron \bgroup \em et al.\egroup }{2023}]{llama2023}
Hugo Touvron, Thibaut Lavril, Gautier Izacard, Xavier Martinet, Marie{-}Anne Lachaux, Timoth{\'{e}}e Lacroix, Baptiste Rozi{\`{e}}re, Naman Goyal, Eric Hambro, Faisal Azhar, Aur{\'{e}}lien Rodriguez, Armand Joulin, Edouard Grave, and Guillaume Lample.
\newblock Llama: Open and efficient foundation language models.
\newblock {\em CoRR}, abs/2302.13971, 2023.

\bibitem[\protect\citeauthoryear{Valenza}{2016}]{triangulation}
Joyce Valenza.
\newblock Truth, truthiness, triangulation: A news literacy toolkit for a “post-truth” world.
\newblock {\em School Library journal}, 2016.

\bibitem[\protect\citeauthoryear{Vrandecic and Kr{\"{o}}tzsch}{2014}]{vrandevcic2014wikidata}
Denny Vrandecic and Markus Kr{\"{o}}tzsch.
\newblock Wikidata: a free collaborative knowledgebase.
\newblock {\em Commun. {ACM}}, 57(10):78--85, 2014.

\bibitem[\protect\citeauthoryear{Wadden \bgroup \em et al.\egroup }{2020a}]{scifact}
David Wadden, Shanchuan Lin, Kyle Lo, Lucy~Lu Wang, Madeleine van Zuylen, Arman Cohan, and Hannaneh Hajishirzi.
\newblock Fact or fiction: Verifying scientific claims.
\newblock In {\em {EMNLP} 2020}, pages 7534--7550, 2020.

\bibitem[\protect\citeauthoryear{Wadden \bgroup \em et al.\egroup }{2020b}]{wadden-etal-2020-fact}
David Wadden, Shanchuan Lin, Kyle Lo, Lucy~Lu Wang, Madeleine van Zuylen, Arman Cohan, and Hannaneh Hajishirzi.
\newblock Fact or fiction: Verifying scientific claims.
\newblock In Bonnie Webber, Trevor Cohn, Yulan He, and Yang Liu, editors, {\em EMNLP}, pages 7534--7550, Online, November 2020.

\bibitem[\protect\citeauthoryear{Wang \bgroup \em et al.\egroup }{2023a}]{fact_survey}
Cunxiang Wang, Xiaoze Liu, Yuanhao Yue, Xiangru Tang, and et~al.
\newblock Survey on factuality in large language models: Knowledge, retrieval and domain-specificity.
\newblock {\em CoRR}, abs/2310.07521, 2023.

\bibitem[\protect\citeauthoryear{Wang \bgroup \em et al.\egroup }{2023b}]{kg-conflict}
Yike Wang, Shangbin Feng, Heng Wang, Weijia Shi, Vidhisha Balachandran, Tianxing He, and Yulia Tsvetkov.
\newblock Resolving knowledge conflicts in large language models.
\newblock {\em CoRR}, abs/2310.00935, 2023.

\bibitem[\protect\citeauthoryear{Yao \bgroup \em et al.\egroup }{2023}]{edit_survey}
Yunzhi Yao, Peng Wang, Bozhong Tian, Siyuan Cheng, Zhoubo Li, Shumin Deng, Huajun Chen, and Ningyu Zhang.
\newblock Editing large language models: Problems, methods, and opportunities.
\newblock {\em EMNLP 2023}, abs/2305.13172, 2023.

\bibitem[\protect\citeauthoryear{Yin \bgroup \em et al.\egroup }{2023}]{Know2023}
Zhangyue Yin, Qiushi Sun, Qipeng Guo, Jiawen Wu, Xipeng Qiu, and Xuanjing Huang.
\newblock Do large language models know what they don't know?
\newblock {\em CoRR}, abs/2305.18153, 2023.

\bibitem[\protect\citeauthoryear{Zhang \bgroup \em et al.\egroup }{2023}]{suvery-Siren}
Yue Zhang, Yafu Li, Leyang Cui, Deng Cai, Lemao Liu, Tingchen Fu, Xinting Huang, Enbo Zhao, Yu~Zhang, Yulong Chen, Longyue Wang, Anh~Tuan Luu, Wei Bi, Freda Shi, and Shuming Shi.
\newblock Siren's song in the {AI} ocean: {A} survey on hallucination in large language models.
\newblock {\em CoRR}, abs/2309.01219, 2023.

\bibitem[\protect\citeauthoryear{Zhao \bgroup \em et al.\egroup }{2023}]{LLM-survey}
Wayne~Xin Zhao, Kun Zhou, and et~al.
\newblock A survey of large language models.
\newblock {\em CoRR}, abs/2303.18223, 2023.

\end{thebibliography}

\end{document}